\documentclass[letterpaper, 10 pt, conference]{ieeeconf}  % Comment this line out if you need a4paper

\IEEEoverridecommandlockouts                              % This command is only needed if 
\makeatletter
\let\NAT@parse\undefined
\makeatother
\usepackage{capt-of} 
\usepackage{amsmath,amsfonts}
\usepackage{array}
\usepackage[caption=false,font=normalsize,labelfont=sf,textfont=sf]{subfig}
\usepackage{textcomp}
\usepackage{stfloats}
\usepackage{url}
\usepackage{verbatim}
\usepackage{graphicx}
\usepackage{cite}
\usepackage{multirow}
\usepackage{xcolor}
\usepackage{url}
\usepackage{colortbl}
\usepackage{booktabs}
\usepackage{makecell}
\usepackage{siunitx}
\usepackage[linesnumbered,ruled,vlined]{algorithm2e}
\usepackage{microtype} 
\usepackage[colorlinks,citecolor=blue]{hyperref}
\usepackage{tabularx}

\title{\LARGE \bf
Towards Path-Creative Navigation: Robot Navigation through Embodied Interaction
}

\author{Haoyu Xi$^{1,2}$, 
Siwei Cheng$^{2}$,
Xiangyuan Liu$^{3}$,
Brenda Li$^{4}$,
Wei Zhang$^{2*}$%
\thanks{$^{1}$School of Computer Science, Shanghai Jiao Tong University, Shanghai, P. R. China}%
\thanks{$^{2}$College of Information Science and Technology, Eastern Institute of Technology, Ningbo, P. R. China}%
\thanks{$^{3}$School of Mechanical and Power Engineering, Zhengzhou university, Zhengzhou, P.R.China}%
\thanks{$^{4}$University of Waterloo}%
\thanks{*Corresponding author. Email: zhw@eitech.edu.cn.}%
}

\begin{document}

\IEEEaftertitletext{%
  \vspace{-1em} % 缩小作者与 Fig. 1 之间的距离
  \noindent
  \begin{minipage}{0.9\textwidth}
    \centering
    \includegraphics[width=\linewidth]{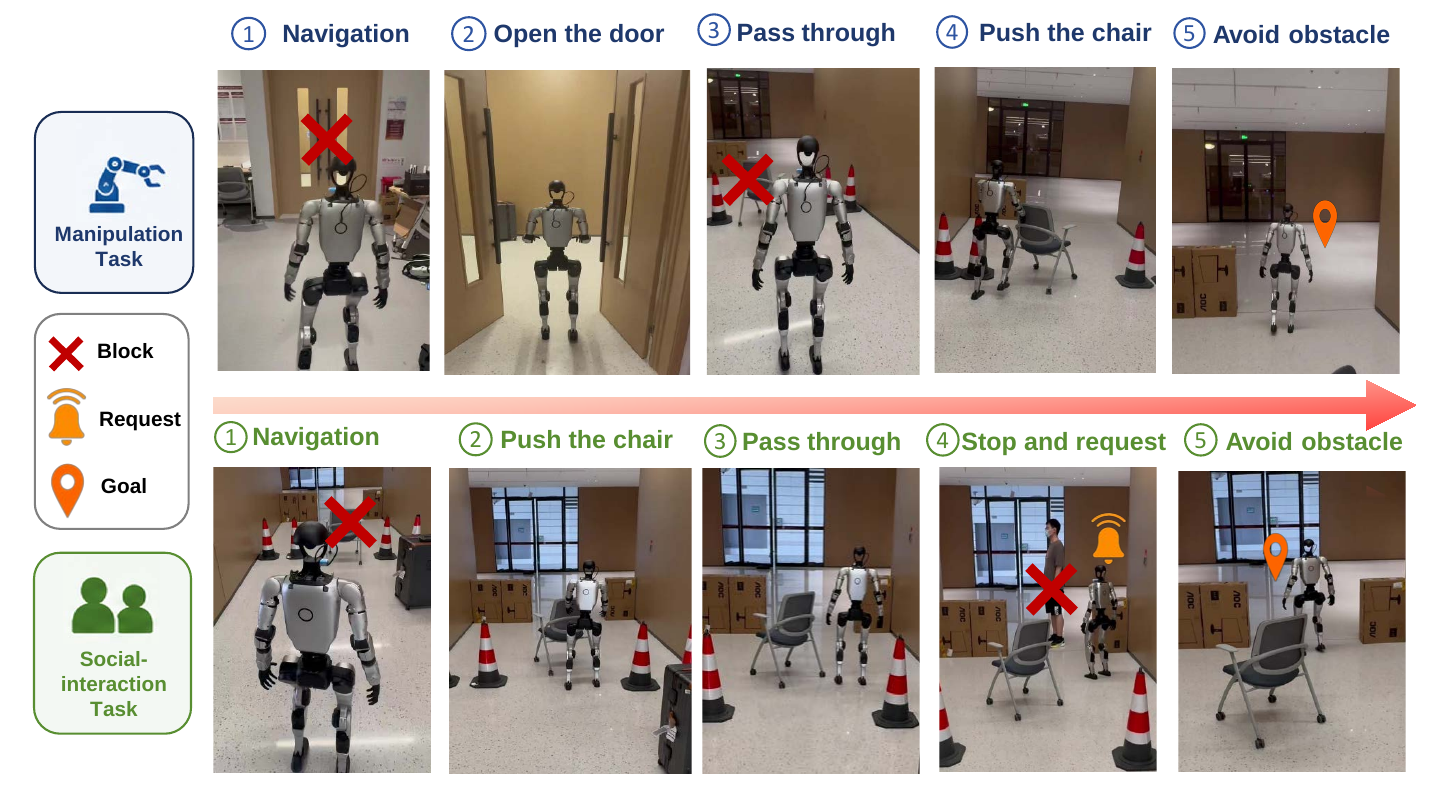}
\captionof{figure}{Real-world experiments on the Unitree G1 humanoid robot. The experiments include manipulation and social-interaction tasks that require the robot to continuously coordinate navigation, obstacle avoidance, and embodied interaction in cluttered environments. Manipulation tasks involve opening doors and pushing movable objects, while social-interaction tasks require the robot to push movable objects and issue a verbal request when a pedestrian blocks a narrow passage. The robot performs tasks using local observations from onboard computer and sensors, including RGB/RGB-D cameras, LiDAR, and odometry, rather than pre-built maps or prior environment knowledge.
}
    \label{fig:overview}
  \end{minipage}
  \vspace{0.5em} % Fig. 1 与摘要之间的距离
}

\maketitle
\thispagestyle{empty}
\pagestyle{empty}

%%%%%%%%%%%%%%%%%%%%%%%%%%%%%%%%%%%%%%%%%%%%%%%%%%%%%%%%%%%%%%%%%%%%%%%%%%%%%%%%
\begin{abstract}

Autonomous navigation in cluttered and constrained environments typically assumes a fixed environment and searches only for paths within existing free space. However, a route can be blocked by an articulated structure, a movable object, or a pedestrian, and reaching the goal may therefore require appropriate embodied interaction with the environment. 
This paper formulates Path-Creative Navigation (PCN), a navigation paradigm in which the robot recovers free space by coordinating locomotion and embodied interaction, and addresses one class of PCN tasks with a mapless framework.
The vision-LiDAR-odometry framework integrates local observations, obstacle geometry, and robot-state estimates through a unified navigation system, providing stable semantic and geometric information for navigation and interaction. 
The traversability-aware decision method utilizes visual and LiDAR distance information to determine whether a blockage is actionable and whether it can be safely bypassed, enabling the robot to interact only when necessary while supporting articulated structure pushing, movable object pushing, obstacle avoidance, and pedestrian requests.
We design simulation scenarios and evaluate different methods in both simulation and two real-world tasks that cannot be completed by conventional navigation alone without embodied interaction. 
These results demonstrate that our framework achieves higher task completion than the baseline methods by coordinating navigation and interaction in cluttered and constrained environments. 

\end{abstract}

%%%%%%%%%%%%%%%%%%%%%%%%%%%%%%%%%%%%%%%%%%%%%%%%%%%%%%%%%%%%%%%%%%%%%%%%%%%%%%%%
\section{INTRODUCTION}

Autonomous navigation enables robots to plan routes and avoid obstacles in cluttered and constrained environments. However, most navigation systems assume a fixed environment and search only for paths within existing free space. In homes, offices, or warehouses, a route may instead be blocked by an articulated structure, a movable object, or a pedestrian in a narrow passage. In such cases, reaching the goal may require changing the environment rather than searching for another collision-free path. The key challenge is therefore not only how to interact with the environment, but when interaction is actually necessary for navigation.

This challenge presents three requirements, particularly in cluttered and constrained environments where free space is limited. First, the robot must determine whether an object truly blocks the route or can be safely bypassed. Interacting with every detected object is inefficient and unsafe, whereas treating all obstacles as immutable may leave a recoverable route blocked.
Second, potential blockers impose different constraints: fixed obstacles require avoidance, movable objects may be pushed, articulated doors must be operated according to their constrained motion, and pedestrians may require a non-contact request. 
Third, interaction must remain tightly coupled with the navigation objective as the scene changes. After interacting with a blocker, the robot must reassess whether the passage is traversable and resume navigation when possible. 
Consequently, the task requires geometric reasoning about route blockage and bypass feasibility, semantic reasoning about potentially actionable objects, and closed-loop interaction as the scene changes.

Existing approaches address individual aspects of this challenge but may not fully satisfy the three requirements.
Conventional navigation cannot recover a route that requires an environmental change. Navigation among movable obstacles (NAMO) \cite{1442130} allows a robot to move obstacles and clear free space, while interactive navigation \cite{8954627} permits physical interaction to support goal reaching. These methods focus on movable-object interaction or predefined manipulation tasks. 
Map-based methods \cite{11266927, 11246770} rely on pre-built maps for global planning, but maps provide incomplete path information when traversable regions are not explicitly represented or are only partially observable. LLM-based methods \cite{11204512,11313696} use semantic reasoning to determine whether and how to interact with objects and plan feasible routes, but LLM-based reasoning adds communication latency and computational overhead. 
Task-and-Motion Planning (TAMP) \cite{084139} integrates navigation and manipulation through explicit action sequences, typically involving higher-level task planning. In parallel, humanoid loco-manipulation methods \cite{11399875,he2026ultra} provide the physical skills required for door opening and object pushing. This paper focuses on the complementary navigation-level problem of determining when interaction is required to recover free space and continue toward the goal.

To address this problem, we formulate Path-Creative Navigation (PCN), which explicitly treats the restoration of traversability as part of the navigation problem itself, and jointly reasons about whether the robot should navigate within the current free space or change the environment to make a route traversable.
PCN covers a broad range of tasks, and this paper addresses one class of them, in which the blockage comes from a movable object, an articulated structure, or a pedestrian. For this class, we present a mapless humanoid navigation framework that integrates vision, LiDAR, and odometry. Visual perception provides object semantics and distance estimates, while LiDAR and odometry provide local geometry and robot-pose information.
A traversability-aware decision method jointly reasons about object semantics, route blockage, and bypass feasibility.
The decision uses only local observations, without a pre-built map or a language model in
the loop.

In summary, the main contributions of this paper are as follows:

\begin{itemize}

\item We formulate Path-Creative Navigation (PCN), a navigation paradigm that reasons over locomotion and actionable embodied interaction to recover free space across different types of route blockages.

\item We propose a traversability-aware decision method that considers local observations to determine whether a blockage is actionable and whether it can be safely bypassed, triggering interaction only when necessary to restore route traversability.

\item We design a vision-LiDAR-odometry fusion framework that combines object semantics, local obstacle geometry, and robot-state estimates within a unified navigation system, supporting articulated structure pushing, movable object pushing, pedestrian request, and navigation toward the goal.

\item We evaluate our framework in both simulation and real-world experiments, demonstrating higher task completion and more reliable interaction selection while coordinating navigation and interaction.

\end{itemize}

\section{RELATED WORK}

\subsection{Robot Mobile Manipulation}

Navigation Among Movable Obstacles (NAMO) relaxes the assumption that the environment remains fixed during navigation. Affordance-based NAMO \cite{9341337}  estimates whether an unknown obstacle can be moved and uses contact-aware planning to complete an otherwise infeasible navigation task. Social concerns also apply when a robot changes the environment. Socially aware NAMO \cite{9340892} adds a placement cost so that a displaced object is less likely to block a corridor, doorway, or other space needed by people. These methods mainly reason about which objects should be moved and where they should be placed. In contrast, our framework does not perform global planning or optimize final object placements; it uses local semantic and geometric observations around the robot-to-goal corridor to assess local traversability and coordinate predefined responses to doors, movable objects, fixed obstacles, and pedestrians within a unified humanoid navigation system.

Interactive navigation permits physical interaction to support goal reaching. Recent methods mainly improve the decision and control required for path clearing.  Yang et al.  \cite{11245985} use hierarchical policy learning to estimate obstacle properties  and generate pushing actions for a mobile manipulator. Dai et al. \cite{11266927} combine adaptive object-dynamics learning with non-prehensile mobile manipulation. For legged manipulators, Bi et al. \cite{11246770} learn an arm-pushing controller that moves large obstacles in narrow spaces.  Manipulate-to-Navigate \cite{11247222} combines visual affordances and manipulability priors to learn feasible end-effector poses for Reach and Door tasks. These methods show that interaction can make a blocked route feasible. However, most of these learning-based approaches train policies or object models for specific interaction tasks and settings. Our framework applies to new visual instances within supported categories, such as doors and boxes with different colors and appearances, without fine-tuning or retraining.

Task-and-Motion Planning (TAMP) connects discrete action choices with continuous geometric and kinematic feasibility. 
A major challenge is the large search space of long-horizon tasks. VLM-TAMP \cite{11128705} addresses this problem by using a vision-language model to generate intermediate subgoals, which divide a long manipulation task into shorter TAMP problems. When a subgoal cannot be refined into a feasible motion, the system queries the model again and replans. Sung et al. \cite{pmlr-v205-sung23a} improve long-horizon TAMP through learned backjumping heuristics. Unlike general TAMP, our framework does not search for or revise a long-horizon sequence of task and motion actions; it makes local traversability decisions.

\subsection{Robot Loco-manipulation}

Loco-manipulation provides the physical skills needed to execute 
such interactions. Recent work improves whole-body safety for mobile manipulators in cluttered environments \cite{10924710}. 
DemoHLM \cite{11399875} generates diverse synthetic trajectories from one simulation demonstration for each task and uses them to train high-level manipulation policies on top of a universal whole-body controller.  ULTRA \cite{he2026ultra} combines physics-driven motion retargeting with a unified  controller that accepts either dense motion references or sparse task goals. 
Our framework performs local observations reasoning from semantic and geometric information around the robot-to-goal corridor, coordinating predefined responses to different blocker types.

Door opening is a related mobility skill, but a door is an articulated structure
rather than a freely movable rigid object. Its motion is constrained by a 
 hinge, and the robot must coordinate the arm, body, and door state. Jang et al. \cite{10132587}
formulate approaching, opening, traversing, and closing as a coupled planning 
problem for a mobile manipulator. Zhang et al.  \cite{zhang2024door} learn a 
policy that opens and traverses both push and pull doors with a legged 
manipulator. These works demonstrate that changing the 
state of an articulated structure may be necessary for mobility. 
In our work, however, opening the door is not the task goal itself. 
The action is selected only when the current door state blocks goal 
reaching and opening it can create sufficient free space.

\section{PROBLEM FORMULATION}
\label{sec:problem}

Consider a mobile embodied robot operating in a potentially changeable
environment $\mathcal{E}_{\mathrm{t}}$ at time step $t$. The robot has an initial
state $x_0 \in \mathcal{X}$ and a navigation goal
$x_{\mathrm{g}} \in \mathcal{X}$. The environment contains a set of elements
\begin{equation}
\mathcal{O}
=
\mathcal{O}_{\mathrm{fix}}
\cup
\mathcal{O}_{\mathrm{mov}}
\cup
\mathcal{O}_{\mathrm{art}},
\end{equation}
where $\mathcal{O}_{\mathrm{fix}}$ denotes fixed obstacles whose states
cannot be deliberately changed by the robot,
$\mathcal{O}_{\mathrm{mov}}$ denotes movable objects whose positions may
change, and $\mathcal{O}_{\mathrm{art}}$ denotes articulated structures
whose configurations can be altered through interaction.
In particular, pedestrians are treated as a special class of movable
objects whose states may change in response to the robot.

At time step $t$, the robot receives an observation
$\mathbf{obs}_{\mathrm{t}}$ of the current environment and selects an
action from
\begin{equation}
\mathcal{A}
=
\mathcal{A}_{\mathrm{nav}}
\cup
\mathcal{A}_{\mathrm{int}},
\end{equation}
where $\mathcal{A}_{\mathrm{nav}}$ denotes locomotion actions that move
the robot through the free space, and
$\mathcal{A}_{\mathrm{int}}$ denotes embodied interactions that may alter
the state of changeable elements in the environment.

An interaction action is represented as
\begin{equation}
a_{\mathrm{t}}^{\mathrm{int}}
=
(o_{\mathrm{t}},m_{\mathrm{t}}),
\qquad
o_{\mathrm{t}} \in
\mathcal{O}_{\mathrm{mov}}
\cup
\mathcal{O}_{\mathrm{art}},
\qquad
m_{\mathrm{t}} \in \mathcal{M}(o_{\mathrm{t}}),
\end{equation}
where $o_{\mathrm{t}}$ denotes the interaction target and
$\mathcal{M}(o_{\mathrm{t}})$ denotes the set of admissible interaction modes for
that target. Depending on the target, an interaction may be physical,
such as changing the position of a movable object or the configuration
of an articulated structure, or non-contact, such as requesting a
pedestrian to clear the passage.

The coupled robot--environment evolution is described by
$x_{\mathrm{t+1}} = f(x_{\mathrm{t}},a_{\mathrm{t}},\mathcal{E}_{\mathrm{t}})$, and $\mathcal{E}_{\mathrm{t+1}} = g(\mathcal{E}_{\mathrm{t}},a_{\mathrm{t}})$.
Navigation actions primarily change the robot state, whereas interaction
actions may additionally change the environment state and thereby alter
the free space available to the robot.

We formulate Path-Creative Navigation as finding a policy
\begin{equation}
f_{\mathrm{\pi}}:
(\mathbf{obs}_{\mathrm{t}},x_{\mathrm{t}},x_{\mathrm{g}})
\mapsto
a_{\mathrm{t}},
\qquad
a_{\mathrm{t}} \in
\mathcal{A}_{\mathrm{nav}}
\cup
\mathcal{A}_{\mathrm{int}},
\end{equation}
such that the resulting robot--environment evolution satisfies the robot
dynamics, collision-avoidance requirements, and interaction constraints,
and eventually reaches the goal region $x_{\mathrm{T}} \in \mathcal{G}(x_{\mathrm{g}})$,
where $\mathcal{G}(x_{\mathrm{g}})$ denotes an admissible goal region around
$x_{\mathrm{g}}$.

The formulation above covers a broad range of environment-changing interactions.  The
actionable elements belong to known categories. Whether an element blocks the route
and whether it can be bypassed are decided from the current observation
$\mathbf{obs}_{\mathrm{t}}$, without a pre-built map or global planning. Interacting with blocking elements is enough to make the passage traversable.

\section{METHODOLOGY}
\label{sec:method}

\begin{figure*}[!t]
	\centering
	\includegraphics[width=0.9\textwidth]{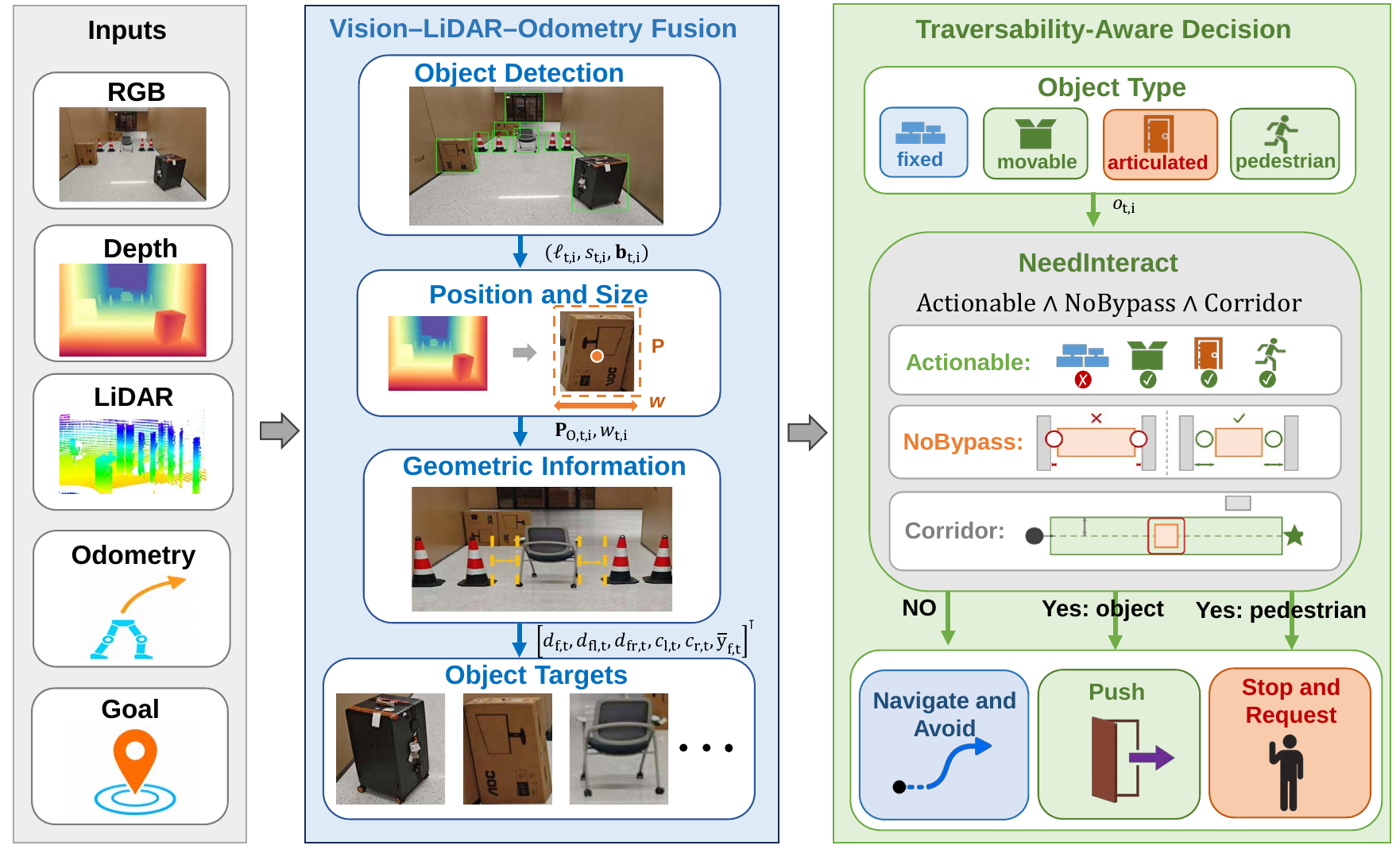}\\
\caption{Overview of the framework. Vision, depth, LiDAR, and odometry are integrated to obtain stable object targets $o_{\mathrm{t,i}}$ and local geometric observations for closed-loop interaction. The traversability-aware decision method jointly evaluates object actionability, $\operatorname{Corridor}(o_{\mathrm{t,i}})$, and $\operatorname{NoBypass}(o_{\mathrm{t,i}})$ to determine $\operatorname{NeedInteract}(o_{\mathrm{t,i}})$. The robot therefore retains navigation for fixed or bypassable obstacles and invokes the corresponding embodied interaction only when necessary. After the scene changes, traversability is reassessed and navigation resumes toward the goal.}
\label{fig:fig2}
\end{figure*}

\subsection{Overview of Our Framework}

We instantiate the class of PCN tasks defined in Section~\ref{sec:problem} with two
modules.  The vision-LiDAR-odometry fusion framework provides stable
semantic targets and local geometric information. The
traversability-aware decision method determines whether the robot
should continue normal navigation, avoid an object, or interact with
an actionable object. The whole framework is shown in Fig.~\ref{fig:fig2}.

\subsection{Vision-LiDAR-Odometry Fusion}
\label{subsec:fusion}

The fusion framework combines visual semantics and depth with LiDAR
geometry and odometry. Vision and depth provide semantic object targets
in the odometry frame, while LiDAR and odometry provide local geometric
information for navigation and interaction.

\subsubsection{Semantic Target Detection}

At time step $t$, the observation $\mathbf{obs}_{\mathrm{t}}$ is instantiated by the RGB image $I_{\mathrm{t}}$, aligned
depth image $D_{\mathrm{t}}$,  LiDAR point cloud $\mathcal{L}_{\mathrm{t}}$,  robot state
$x_{\mathrm{t}}$, and  final navigation goal $x_{\mathrm{g}}$. We use YOLO-World~\cite{10657649},
a real-time open-vocabulary object detector, to process $I_{\mathrm{t}}$ with text
prompts. For the $i$-th detection, the detector outputs the semantic label $\ell_{\mathrm{t,i}}$, confidence
score $s_{\mathrm{t,i}}$, and bounding box:
\begin{equation}
\mathbf{b}_{\mathrm{t,i}}
=
[u_{\mathrm{1,t,i}},v_{\mathrm{1,t,i}},u_{\mathrm{2,t,i}},v_{\mathrm{2,t,i}}]^{\top},
\label{eq:detection_output}
\end{equation}
where $\mathbf{b}_{\mathrm{t,i}}$ is the bounding box with top-left corner
$(u_{\mathrm{1,t,i}},v_{\mathrm{1,t,i}})$ and bottom-right corner
$(u_{\mathrm{2,t,i}},v_{\mathrm{2,t,i}})$.
Labels of articulated structures, such as doors, are mapped to
$\mathcal{O}_{\mathrm{art}}$, while movable objects, such as boxes,
cartons, cubes, and chairs, are mapped to $\mathcal{O}_{\mathrm{mov}}$.
Person-related labels are also mapped to $\mathcal{O}_{\mathrm{mov}}$,
but are handled separately by the pedestrian interaction strategy.
Object size is used for geometric reasoning rather than semantic
classification.

\subsubsection{Depth-Assisted Size Estimation}

The RGB bounding box is mapped to the aligned depth image, where a
central region is selected to reduce contamination from background
pixels and bounding-box edges. Invalid depth measurements are removed.
Let $\mathcal{V}_{\mathrm{t,i}}$ denote the valid depth samples in this region.
We retain a fraction $\eta$ of the nearest samples and use their median
as the object depth:
\begin{equation}
z_{\mathrm{t,i}}
=
\operatorname{median}
\left(
\operatorname{Near}_{\mathrm{\eta}}
\left(
\mathcal{V}_{\mathrm{t,i}}
\right)
\right).
\label{eq:nearest_depth}
\end{equation}
This provides a more robust depth estimate than using a single center
pixel.

Let $(u_{\mathrm{t,i}},v_{\mathrm{t,i}})$ denote the center of the selected depth region.
The corresponding 3-D point in the camera frame is obtained by
back-projection:
\begin{equation}
\mathbf{p}_{\mathrm{C,t,i}}
=
\begin{bmatrix}
\dfrac{(u_{\mathrm{t,i}}-c_{\mathrm{x}})z_{\mathrm{t,i}}}{f_{\mathrm{x}}}\\[2mm]
\dfrac{(v_{\mathrm{t,i}}-c_{\mathrm{y}})z_{\mathrm{t,i}}}{f_{\mathrm{y}}}\\[2mm]
z_{\mathrm{t,i}}
\end{bmatrix},
\label{eq:target_camera}
\end{equation}
where $f_{\mathrm{x}}$ and $f_{\mathrm{y}}$ are the camera focal lengths, and $c_{\mathrm{x}}$ and $c_{\mathrm{y}}$
are the principal-point coordinates.

The estimated 3-D position in the odometry frame $\mathbf{p}_{\mathrm{O,t,i}}$ is transformed as:
\begin{equation}
\begin{bmatrix}
\mathbf{p}_{\mathrm{O,t,i}}\\
1
\end{bmatrix}
=
\mathbf{T}_{\mathrm{OB}}(x_{\mathrm{t}})
\mathbf{T}_{\mathrm{BC}}
\begin{bmatrix}
\mathbf{p}_{\mathrm{C,t,i}}\\
1
\end{bmatrix},
\label{eq:target_odom}
\end{equation}
where $\mathbf{T}_{\mathrm{BC}}\in SE(3)$ is the fixed camera-to-base
transformation and $\mathbf{T}_{\mathrm{OB}}(x_{\mathrm{t}})\in SE(3)$ is the
base-to-odometry transformation obtained from $x_{\mathrm{t}}$.

The metric width of the detected object is approximated by
\begin{equation}
w_{\mathrm{t,i}}
=
\frac{(u_{\mathrm{2,t,i}}-u_{\mathrm{1,t,i}})z_{\mathrm{t,i}}}{f_{\mathrm{x}}},
\label{eq:metric_object_width}
\end{equation}
and is used for geometric reasoning and task-corridor construction.
A target is accepted after $N_{\mathrm{s}}$ consecutive observations are spatially
consistent. Their average position is used to update $\mathbf p_{O,t,i}$, which is subsequently treated as the stable target position and locked once the robot enters the close-range interaction region.

For each accepted detection, the stable object interaction target is defined as
\begin{equation}
{o}_{\mathrm{t,i}}
=
\left(
\ell_{\mathrm{t,i}},\,
\mathbf{p}_{\mathrm{O,t,i}},\,
w_{\mathrm{t,i}}
\right),
\label{eq:stable_object_target}
\end{equation}
which combines semantic information with the metric attributes required
for subsequent traversability and interaction reasoning. The selected candidate ${o}_{\mathrm{t,i}}$ is used as the interaction target ${o}_{\mathrm{t}}$.

\subsubsection{Local Geometric Information}

LiDAR points are transformed into the robot frame using the current
odometry state. After height, distance, and range filtering, the point
cloud is divided into frontal and lateral regions. The local geometric
observation is summarized as
\begin{equation}
\mathbf{g}_{\mathrm{t}}
=
\left[
d_{\mathrm{f,t}},\,
d_{\mathrm{fl,t}},\,
d_{\mathrm{fr,t}},\,
c_{\mathrm{l,t}},\,
c_{\mathrm{r,t}},\,
\bar{y}_{\mathrm{f,t}}
\right]^{\top},
\label{eq:lidar_summary}
\end{equation}
where $d_{\mathrm{f,t}}$ is the front distance, $d_{\mathrm{fl,t}}$ and $d_{\mathrm{fr,t}}$ are
the front-left and front-right distances, $c_{\mathrm{l,t}}$ and $c_{\mathrm{r,t}}$ are
the scene-level left and right clearances, and $\bar{y}_{\mathrm{f,t}}$ is the lateral
position of the main frontal structure.
The fusion framework outputs the object targets $o_{\mathrm{t,i}}$ in the
odometry frame together with the local geometric observation
for subsequent traversability-aware decision making.

\subsection{Traversability-Aware Decision Method}
\label{subsec:decision}

The traversability-aware decision method determines whether a detected
object should be handled by navigation or embodied interaction. The
decision jointly considers the object's semantic category, its relation
to the task corridor, and the available LiDAR clearance. Fixed obstacles
are handled by navigation and obstacle avoidance, whereas movable objects
and articulated structures are considered for interaction only when they
block the active route and cannot be safely bypassed.

\subsubsection{Task Corridor and Bypass Evaluation}

Let $\mathbf{p}_{\mathrm{t}}^{\mathrm{r}}=\pi_{\mathrm{xy}}(x_{\mathrm{t}})$ and
$\mathbf{p}_{\mathrm{g}}=\pi_{\mathrm{xy}}(x_{\mathrm{g}})$ denote the horizontal positions of the
robot and the final goal, respectively. The longitudinal and lateral
directions of the task corridor are defined as
\begin{equation}
\mathbf{e}_{\mathrm{g,t}}
=
\frac{\mathbf{p}_{\mathrm{g}}-\mathbf{p}_{\mathrm{t}}^{\mathrm{r}}}
{\|\mathbf{p}_{\mathrm{g}}-\mathbf{p}_{\mathrm{t}}^{\mathrm{r}}\|},
\qquad
\mathbf{e}_{\mathrm{\perp,t}}
=
\begin{bmatrix}
e_{\mathrm{g,t}}^{\mathrm{y}} & -e_{\mathrm{g,t}}^{\mathrm{x}}
\end{bmatrix}^{\top}.
\label{eq:task_directions}
\end{equation}

Let
$\bar{\mathbf{p}}_{\mathrm{t,i}}
=[p_{\mathrm{O,t,i}}^{\mathrm{x}},p_{\mathrm{O,t,i}}^{\mathrm{y}}]^{\top}$
denote the horizontal position of the detected object $o_{\mathrm{t,i}}$.
An object overlaps the task corridor if it lies in front of the robot
and its lateral extent intersects the required passage:
\begin{equation}
\begin{aligned}
\operatorname{Corridor}(o_{\mathrm{t,i}})
={}&
\left[
(\bar{\mathbf{p}}_{\mathrm{t,i}}-\mathbf{p}_{\mathrm{t}}^{\mathrm{r}})^{\top}
\mathbf{e}_{\mathrm{g,t}}>0
\right]
\\
&\land
\left[
\left|
(\bar{\mathbf{p}}_{\mathrm{t,i}}-\mathbf{p}_{\mathrm{t}}^{\mathrm{r}})^{\top}
\mathbf{e}_{\mathrm{\perp,t}}
\right|
\leq
r_{\mathrm{r}}+\frac{w_{\mathrm{t,i}}}{2}+m_{\mathrm{c}}
\right],
\end{aligned}
\label{eq:corridor_condition}
\end{equation}
where $r_{\mathrm{r}}$ is the robot half-width, $w_{\mathrm{t,i}}$ is the estimated object
width, and $m_{\mathrm{c}}$ is a safety margin.

For each candidate object, LiDAR points are selected within a
target-centered region aligned with the current robot heading.
After excluding the object's lateral extent, supported LiDAR clusters
on its two sides provide the object-specific clearances
$c_{\mathrm{l,t,i}}$ and $c_{\mathrm{r,t,i}}$.
The object is considered non-bypassable when both clearances are no
greater than the threshold $c_{\mathrm{th}}$:
\begin{equation}
\operatorname{NoBypass}(o_{\mathrm{t,i}})
=
\left(c_{\mathrm{l,t,i}}\leq c_{\mathrm{th}}\right)
\land
\left(c_{\mathrm{r,t,i}}\leq c_{\mathrm{th}}\right).
\label{eq:no_bypass_condition}
\end{equation}
This condition is evaluated only when both sides contain sufficiently
supported and laterally extended LiDAR observations, and it must remain
valid for several consecutive scans. If either side provides sufficient
clearance or the LiDAR evidence is unreliable, the robot continues
navigation without triggering interaction.

\subsubsection{Interaction Necessity}

The semantic label $\ell_{\mathrm{t,i}}$ determines whether  the target $o_{\mathrm{t,i}}$ belongs to the fixed, movable, or articulated object set. An object is actionable if it is movable or articulated:
\begin{equation}
\operatorname{Actionable}(o_{\mathrm{t,i}})
=
\left(
o_{\mathrm{t,i}} \in
\mathcal{O}_{\mathrm{mov}}
\cup
\mathcal{O}_{\mathrm{art}}
\right).
\label{eq:actionable_object}
\end{equation}

Interaction is triggered only when an actionable object overlaps the
task corridor and cannot be safely bypassed:
\begin{equation}
\begin{aligned}
\operatorname{NeedInteract}(o_{\mathrm{t,i}})
={}&
\operatorname{Actionable}(o_{\mathrm{t,i}})
\\
&\land
\operatorname{Corridor}(o_{\mathrm{t,i}})
\land
\operatorname{NoBypass}(o_{\mathrm{t,i}}).
\end{aligned}
\label{eq:need_interaction}
\end{equation}

If $\operatorname{NeedInteract}(o_{\mathrm{t,i}})=0$, the navigation action
$a_{\mathrm{t}}^{\mathrm{nav}}$ is retained. Otherwise, the candidate is selected
as the interaction target, and an
admissible mode $m_{\mathrm{t}}$ is executed through $a_{\mathrm{t}}^{\mathrm{int}}$.

\subsection{Embodied Interaction Execution}

When $\operatorname{NeedInteract}(o_{\mathrm{t,i}})=1$, the selected object is
used as the interaction target $o_{\mathrm{t}}$, and an admissible interaction mode
$m_{\mathrm{t}} \in \mathcal{M}(o_{\mathrm{t}})$ is executed through
$a_{\mathrm{t}}^{\mathrm{int}}=(o_{\mathrm{t}},m_{\mathrm{t}})$. The execution strategy depends on the
semantic category of the target. Fixed obstacles remain handled by
navigation and obstacle avoidance.

\begin{itemize}

\item \textbf{Physical interaction:}
For movable objects and articulated structures, the robot first
approaches the selected target and aligns its body with the interaction
direction. It then moves its arms to the required pose and executes the
corresponding physical interaction. $\mathbf{g}_{\mathrm{t}}$  provides
closed-loop feedback for approach, body alignment, speed adjustment,
and interaction completion. Once the passage becomes traversable, the
robot retracts its arms and resumes navigation. If the interaction
cannot be completed safely, the robot moves away from the target and
repeats the approach and alignment procedure.

\item \textbf{Pedestrian request:}
Pedestrians, treated as a special class of movable objects, are handled
through non-contact interaction. When a pedestrian blocks the active
route and satisfies $\operatorname{NeedInteract}(o_{\mathrm{t,i}})=1$, the robot
stops and issues a verbal request to clear the passage. Navigation
resumes once the passage becomes traversable.

\end{itemize}

Our framework integrates DRL-DCLP \cite{10900448} as the low-level point-goal navigation module for obstacle avoidance.
The illustration of  embodied interaction is shown in Fig.~\ref{fig:action} in Section V. Both simulation and real-world tests use the same interaction executions.

\begin{figure}[t]
	\centering
	\includegraphics[width=0.9\columnwidth]{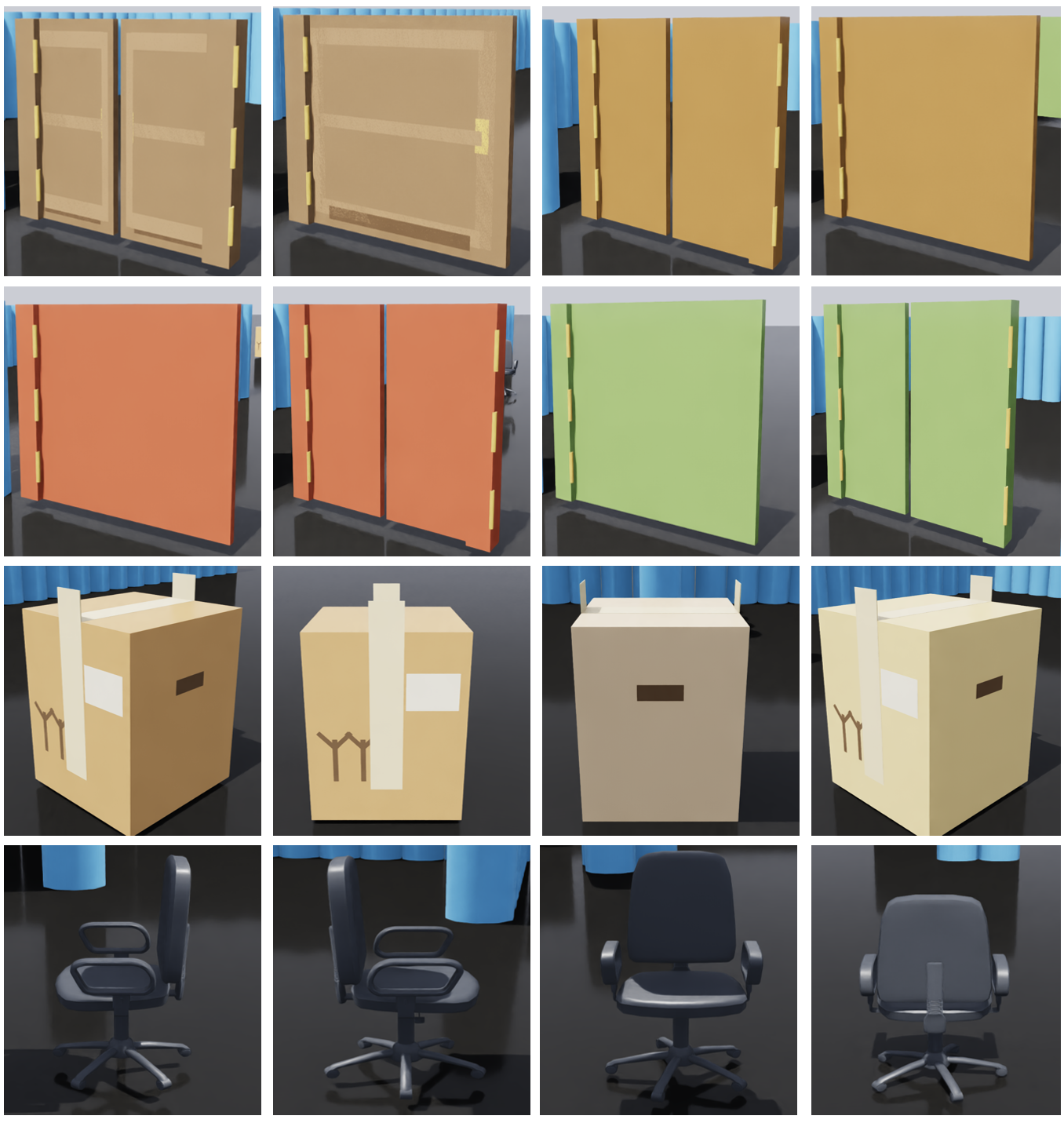}\\
	\caption{ Objects in simulation. The simulated environments include doors of different sizes, colors, and types; boxes of varying colors and sizes; and chairs with different sizes and orientations.
	}\label{fig:env}
\end{figure}

\begin{figure}[t]
	\centering
	\includegraphics[width=0.9\columnwidth]{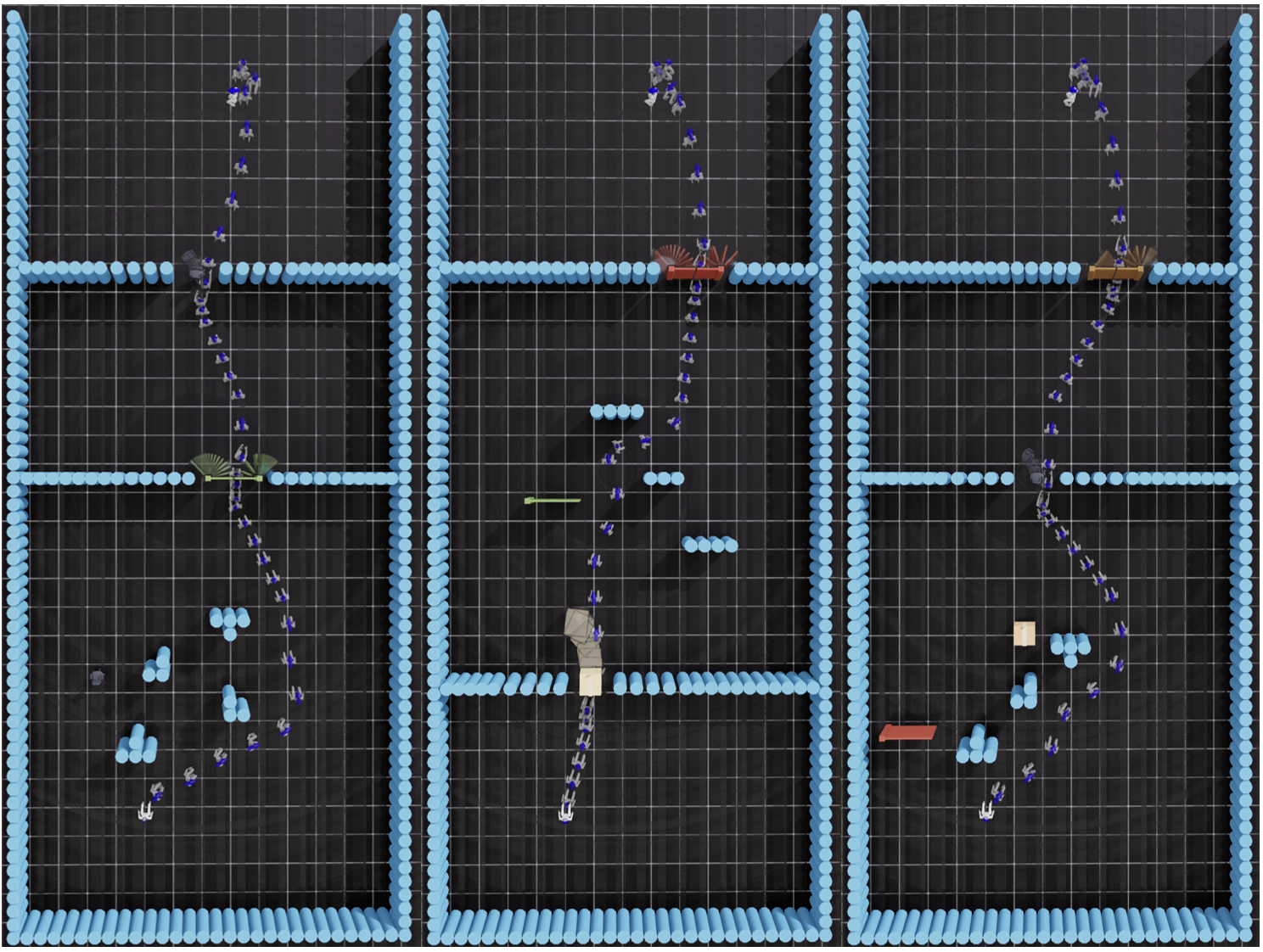}\\
	\caption{ Experimental trajectories in simulation. Our method successfully performs the necessary interactions while treating non-blocking actionable objects as normal obstacles, avoiding unnecessary interactions and reaching the goal.
	}\label{fig:simu}
\end{figure}

\section{EXPERIMENTS}
\subsection{Simulation Experiments Setup and Results Analysis}

To evaluate our method, we construct 70 scenarios in NVIDIA Isaac Sim. As shown in Fig.~\ref{fig:env}, each scenario contains fixed obstacles, movable objects, and articulated structures with varied placements. Only a subset of the detected doors, chairs, and boxes actually blocks the route, requiring the robot to distinguish interaction-relevant obstacles from non-blocking objects that should be handled by normal navigation and obstacle avoidance.
In the Isaac Sim simulation, we use a Unitree G1 humanoid robot equipped with a Unitree L1 LiDAR and an Intel RealSense D435i depth camera. The experiments are conducted on a system equipped with an NVIDIA GeForce RTX 5090 GPU, with Robot Operating System (ROS) for communication.

We compare our proposed method against four methods representing different interaction-selection strategies. DRL-DCLP \cite{10900448} is the navigation-only baseline: it avoids fixed obstacles but cannot modify the environment when the route is blocked. Always-Interact invokes the corresponding interaction for every detected actionable object, regardless of whether the object blocks the route or can be bypassed. Proximity-Trigger initiates an interaction whenever a detected actionable object enters a predefined distance threshold, without explicitly evaluating local bypass feasibility. Oracle-Selective uses the Simulator Ground Truth (GT) to identify the objects whose states must be changed to reach the goal. 
All methods use the embodied interaction introduced in this work for door opening and movable-object pushing. Ours w/o LiDAR removes LiDAR-based geometric information to evaluate its contribution to decision and interaction, while Ours w/o NeedInteract removes the traversability-aware decision module and selects any actionable object detected in front of the robot.
All methods use the same robot, initial state, goal, navigation policy, and evaluation budget. Each method is evaluated twice in every scenario, resulting in 140 trials per method.

In this experiment, SPL measures path efficiency. A selection is a positive decision on a detected target, and an initiation is a selection that enters the interaction stage, regardless of eventual success.
Recall is the fraction of required targets selected at least once, and MIR is the fraction never initiated. Precision is the fraction of selections on required and not previously selected targets, and F1 score is the harmonic mean of precision and recall.
UIR measures initiations involving non-required or previously operated targets divided by all initiations. Recovery retries count once per selection. Mean Time is the mean total time of successful trials.
Required targets are identified from scene ground truth provided by the simulator, and Oracle-Selective is reported separately as a ground truth reference.
Required interactions are identified from scene ground truth provided by the simulator.  Oracle-Selective is reported separately as a ground truth reference. Additionally, we visualize three experimental trajectories in Fig.~\ref{fig:simu}.

\newlength{\twotableheight}
\setlength{\twotableheight}{6.0cm}

\begin{table*}[!t]
\centering

\begin{minipage}[t][\twotableheight][t]{0.655\textwidth}
\centering
\vspace{0pt}

\caption{Quantitative comparison of different methods in simulation}
\label{tab:simulation_results}

\footnotesize
\setlength{\tabcolsep}{3pt}
\renewcommand{\arraystretch}{1.08}

\resizebox{\linewidth}{!}{%
\begin{tabular}{@{}lccccccc@{}}
\toprule
\textbf{Method}
& \makecell{\textbf{Success}\\\textbf{(\%)} $\uparrow$}
& \makecell{\textbf{SPL}\\$\uparrow$}
& \makecell{\textbf{Time}\\\textbf{(s)} $\downarrow$}
& \makecell{\textbf{Recall}\\\textbf{(\%)} $\uparrow$}
& \makecell{\textbf{F1}\\\textbf{(\%)} $\uparrow$}
& \makecell{\textbf{UIR}\\\textbf{(\%)} $\downarrow$}
& \makecell{\textbf{MIR}\\\textbf{(\%)} $\downarrow$} \\
\midrule
DRL-DCLP
& 0.00 & -- & -- & 0.00 & -- & -- & 100.00 \\

Always-Interact
& 45.00 & 0.3936 & 380.09 & 83.57 & 53.00 & 57.41 & 17.86 \\

Proximity-Trigger
& 85.71 & 0.7547 & 327.64 & 92.50 & 91.36 & \textbf{7.86} & 7.86 \\

Ours w/o LiDAR
& 46.43 & 0.4109 & 391.38 & 83.93 & 53.41 & 57.70 & 18.57 \\

Ours w/o NeedInteract
& 47.14 & 0.4290 & 378.11 & 89.29 & 64.60 & 46.26 & 12.86 \\

\textbf{Ours}
& \textbf{93.57} & \textbf{0.8329} & \textbf{317.24}
& \textbf{93.57} & \textbf{91.61} & 9.66 & \textbf{6.43} \\

\midrule
Oracle-Selective (GT)
& 100.00 & 0.9100 & 266.62 & 100.00 & 100.00 & 0.00 & 0.00 \\
\bottomrule
\end{tabular}%
}
\vfill
\end{minipage}
\hfill
% =========================================================
% Real-world results
% =========================================================
\begin{minipage}[t][\twotableheight][t]{0.285\textwidth}
\centering
\vspace{0pt}

\caption{Real-world results.}
\label{tab:real_world_results}

\footnotesize
\setlength{\tabcolsep}{3pt}
\renewcommand{\arraystretch}{1.16}

\resizebox{\linewidth}{!}{%
\begin{tabular}{@{}lccc@{}}
\toprule
\textbf{Method}
& \makecell{\textbf{Success}\\(\%) $\uparrow$}
& \makecell{\textbf{Intervention}\\(\%) $\downarrow$}
& \makecell{\textbf{Time}\\(s) $\downarrow$} \\
\midrule

\multicolumn{4}{c}{\textbf{Manipulation Task}} \\
\cmidrule(lr){1-4}
DRL-DCLP & 0 & 100 & -- \\
Always-Interact & 0 & 100 & -- \\
Proximity-Trigger & 20 & 80 & 91.5 \\
\textbf{Ours} & \textbf{80} & \textbf{20} & \textbf{88.7} \\

\addlinespace[3pt]
\multicolumn{4}{c}{\textbf{Social-interaction Task}} \\
\cmidrule(lr){1-4}
DRL-DCLP & 0 & 100 & -- \\
Always-Interact & 0 & 100 & -- \\
Proximity-Trigger & 10 & 90 & 60.0 \\
\textbf{Ours} & \textbf{70} & \textbf{30} & \textbf{58.5} \\

\bottomrule
\end{tabular}%
}
\vfill
\end{minipage}

\end{table*}

\begin{figure*}[!t]
\centering
\includegraphics[width=\textwidth]{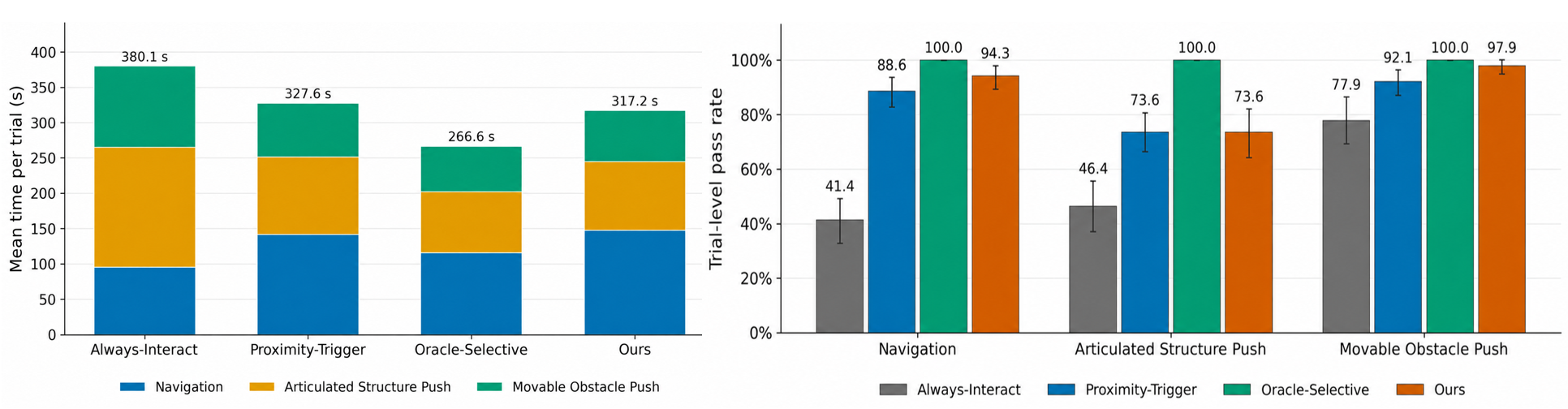}
\caption{Average execution time (left) and trial-level pass rate (right) of the evaluated methods.}
\label{fig:stage_pass}
\end{figure*}

\begin{figure*}[!t]
	\centering
	\includegraphics[width=0.8\textwidth]{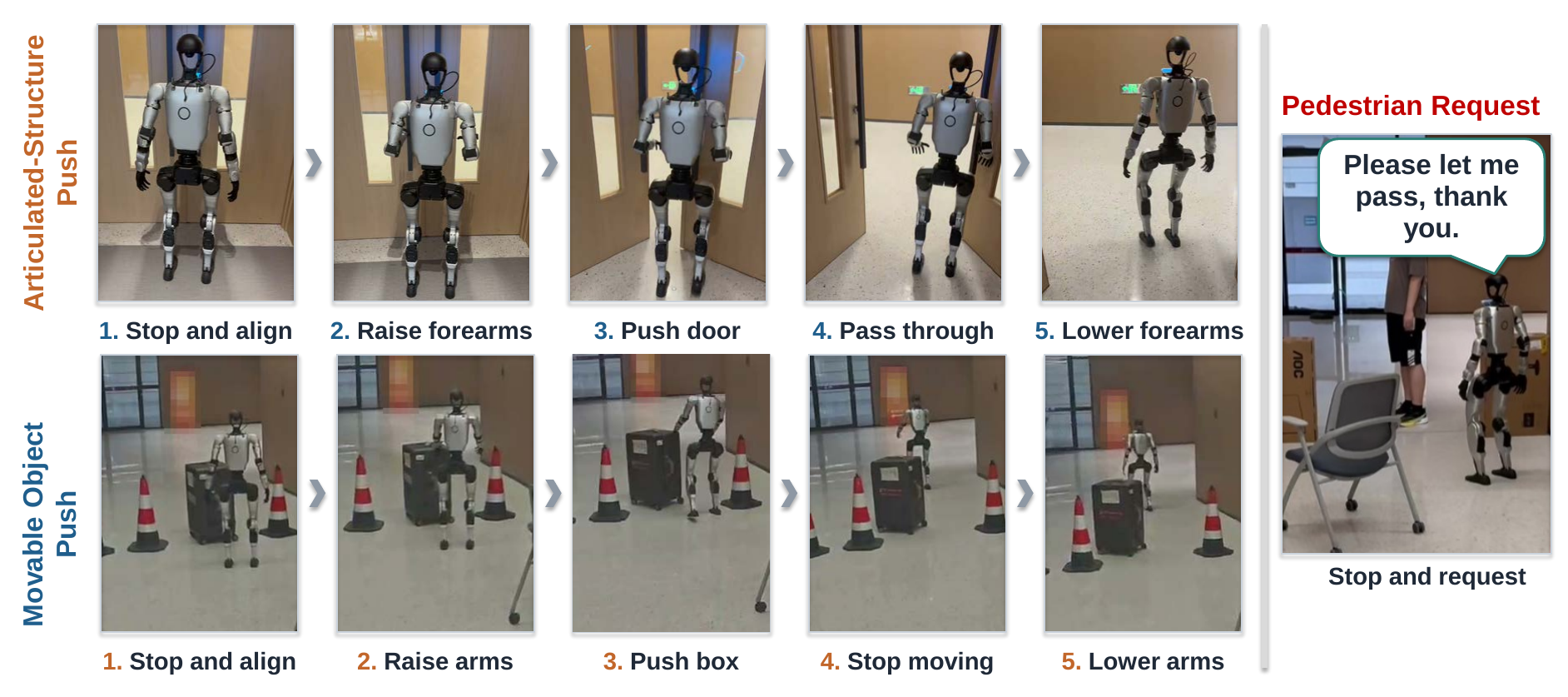}\\
    \caption{Real-world embodied interaction. Interaction executions include articulated structure push and movable obstacle push. The pedestrian request action makes the robot play a verbal request and stop moving until  $\operatorname{NeedInteract}(o_{\mathrm{t,i}})=0$.}
\label{fig:action}
\end{figure*}

Table~\ref{tab:simulation_results} reports the quantitative comparison of different methods in simulation.
$-$ denotes a metric that is not applicable. Among the non-oracle
methods, our work achieves the highest success rate, SPL,
interaction recall, and F1 score. Compared with
Proximity-Trigger, our work improves task completion and reduces the MIR
from 7.86\% to 6.43\%. DRL-DCLP
fails when environmental interaction is required.
Always-Interact executes the same interaction actions as ours but without the decision, and its
UIR of 57.41\% and success rate of 45.00\%.
Oracle-Selective uses ground-truth information provided by the simulator to identify objects that require interaction. With the targets known, our embodied interaction and navigation achieve a 100\% success rate, demonstrating their technical viability.
As shown in Fig.~\ref{fig:stage_pass}, our work achieves trial-level pass rates of 94.3\%, 73.6\%, and 97.9\% for navigation, articulated-structure pushing, and movable-object pushing, respectively.
A trial is counted as successful for a given stage only if it reaches that stage and successfully completes all required steps.

\subsection{Real-World Experiments Setup and Results Analysis}

We further conduct real-world experiments in cluttered and constrained environments. Our experimental platform consists of a Unitree G1 humanoid robot equipped with a Livox Mid-360 LiDAR, an Intel RealSense D435i depth camera, and an NVIDIA Jetson Xavier NX for onboard computation.
Object detection runs at 15 Hz, and planning runs at 10 Hz.
To increase the difficulty beyond simulation, we construct cluttered scenes in narrow corridors with limited free space. The environments contain previously unseen obstacles, including road barriers, suitcases, and chairs, with some objects placed close to the corridor walls. Consequently, unnecessary or incorrect interactions can easily block the remaining free space, cause the robot to become stuck, or cause a collision, requiring intervention. We evaluate two real-world tasks, with 10 trials for each task:

\begin{itemize}

\item Manipulation task: we construct a multi-stage environment
that connects two areas. The robot first pushes open a closed door that
blocks access to the next area. It then continues navigation, avoids
static obstacles, and performs the required interaction tasks. This
setting evaluates whether the robot can maintain goal-directed behavior
across sequential navigation and manipulation steps.

\item Social-interaction task: a pedestrian blocks the path in front
of the robot. The robot stops at a safe distance and provides a verbal
prompt that asks the pedestrian to clear the passage. The robot resumes
navigation only after the passage becomes available. This setting
additionally evaluates whether the robot can restore access through non-contact interaction with a pedestrian. 

\end{itemize}

Key parameters are: $N_{\mathrm{s}} = 3$, $\eta = 0.35$, $r_{\mathrm{r}} = 0.35m$, $m_{\mathrm{c}} = 0.25m$, $c_{\mathrm{th}} = 0.85m$, and the navigation velocity is $0.5 m/s$.
The real-world experiments are illustrated in Fig.~\ref{fig:overview}, and representative interaction executions are shown in Fig.~\ref{fig:action}. As reported in Table~\ref{tab:real_world_results}, PCN achieves success rates of 80\% and 70\% in the manipulation and social-interaction tasks, respectively, whereas DRL-DCLP and Always-Interact fail to complete either task. Proximity-Trigger achieves only 20\% and 10\% success rates.  Its proximity-based interaction strategy is unreliable in narrow corridors, where actionable objects may lie close to the walls without blocking the route, and unnecessary interactions can reduce the remaining free space, cause the robot to become stuck, or result in a collision. In contrast, PCN reduces the intervention rates to 20\% and 30\% by explicitly reasoning about route blockage and bypass feasibility. These results demonstrate the importance of selectively triggering interaction for navigation in constrained real-world environments.

\section{CONCLUSIONS}

This paper formulates Path-Creative Navigation (PCN), a navigation paradigm that restores traversability through embodied interaction. We instantiate PCN with a mapless vision-LiDAR-odometry framework and a traversability-aware decision method that determines whether a blockage is actionable and bypassable. The framework coordinates navigation, obstacle avoidance, articulated-structure interaction, object pushing, and pedestrian requests while maintaining progress toward the goal. Simulation and real-world experiments show higher task completion  than baseline methods, demonstrating that navigation can recover feasible routes by deciding when environmental interaction is necessary.
In future work, we plan to integrate whole-body control and extend PCN to support more challenging navigation and manipulation tasks in complex environments.

\bibliographystyle{IEEEtran}
\bibliography{ref}

\addtolength{\textheight}{-12cm}   % This command serves to balance the column lengths

%%%%%%%%%%%%%%%%%%%%%%%%%%%%%%%%%%%%%%%%%%%%%%%%%%%%%%%%%%%%%%%%%%%%%%%%%%%%%%%%

%%%%%%%%%%%%%%%%%%%%%%%%%%%%%%%%%%%%%%%%%%%%%%%%%%%%%%%%%%%%%%%%%%%%%%%%%%%%%%%%

%%%%%%%%%%%%%%%%%%%%%%%%%%%%%%%%%%%%%%%%%%%%%%%%%%%%%%%%%%%%%%%%%%%%%%%%%%%%%%%%

\end{document}